\documentclass[10pt,twocolumn]{article}
\usepackage[utf8]{inputenc}
\usepackage{mathptmx} 
\usepackage{geometry}
\usepackage{graphicx}
\usepackage{hyperref}
\usepackage{titlesec}
\usepackage{lipsum} 
\usepackage{algorithmic}
\usepackage{tikz}

\usepackage{url}
\usepackage[boxed,ruled]{algorithm2e}

\usepackage{mathtools}

\newcommand{\framework}{VeriPPA\xspace}
\usepackage{wrapfig}
\usepackage{subfig}
\usepackage{pdfpages}
\usepackage{algorithmic}

\titleformat{\section}{\normalfont\Large\bfseries}{\thesection}{1em}{}
\titleformat{\subsection}{\normalfont\large\bfseries}{\thesubsection}{1em}{}
\titleformat{\subsubsection}{\normalfont\normalsize\bfseries}{\thesubsubsection}{1em}{}

\title{Advanced Large Language Model (LLM) - Driven Verilog Development: Enhancing Power, Performance, and Area Optimization in Code Synthesis }

\author{
    Kiran Thorat\textsuperscript{$\ast$}, Jiahui Zhao\textsuperscript{$\ast$}, Yaotian Liu\textsuperscript{$\dagger$}, \\
    Hongwu Peng\textsuperscript{$\ast$}, Xi Xie\textsuperscript{$\ast$}, Bin Lei\textsuperscript{$\ast$}, \\
    Jeff Zhang\textsuperscript{$\dagger$}, Caiwen Ding\textsuperscript{$\ast$}
}

\date{
    \textsuperscript{$\ast$}University of Connecticut, \textsuperscript{$\dagger$}Arizona State University
}

\begin{document}

\maketitle

\begin{abstract}

The increasing use of Advanced Language Models (ALMs) in diverse sectors, particularly due to their impressive capability to generate top-tier content following linguistic instructions, forms the core of this investigation. This study probes into ALMs' deployment in electronic hardware design, with a specific emphasis on the synthesis and enhancement of Verilog programming. We introduce an innovative framework, crafted to assess and amplify ALMs' productivity in this niche. The methodology commences with the initial crafting of Verilog programming via ALMs, succeeded by a distinct dual-stage refinement protocol. The premier stage prioritizes augmenting the code's operational and linguistic precision, while the latter stage is dedicated to aligning the code with Power-Performance-Area (PPA) benchmarks, a pivotal component in proficient hardware design. This bifurcated strategy, merging error remediation with PPA enhancement, has yielded substantial upgrades in the caliber of ALM-created Verilog programming. Our framework achieves an 81.37\% rate in linguistic accuracy and 62.0\% in operational efficacy in programming synthesis, surpassing current leading-edge techniques, such as 73\% in linguistic accuracy and 46\% in operational efficacy. These findings illuminate ALMs' aptitude in tackling complex technical domains and signal a positive shift in the mechanization of hardware design operations. 


\end{abstract}
\section{Introduction}
With Moore's law driving increased design complexity and chip capacity, VLSI design and verification require more effort.
Machine learning (ML)  has successfully integrated into EDA for logic synthesis \cite{8351885, 9045559}, placement \cite{10.1145/2228360.2228497}, routing \cite{10.1145/3372780.3375560, 8533535}, testing \cite{10.1145/2429384.2429404, 10.1145/3194554.3194561, 9000131}, and verification \cite{10.1145/775832.775907, 10.1145/3195970.3196059}. 
The popularity of agile hardware design exploration has been on the rise due to the growth of large language models (LLMs). A promising direction is using natural language instruction to generate hardware description language (HDL). e.g., Verilog, aiming to greatly lower hardware design barriers and increase design productivity, especially for users who do not possess extensive expertise in chip design.
Despite the efforts, Verilog benchmarking has unique challenges in terms of the wide range of hardware designs~\cite{liu2023verilogeval}.

Two orthogonal research and development trends have both attracted enormous interests~\cite{thakur2023benchmarking, lu2023rtllm,liu2023verilogeval,blocklove2023chip,chang2023chipgpt}. The \textit{first trend} is efficiently finetuning LLMs such as CodeGen~\cite{nijkamp2022codegen}, with representatives works such as Thakur \textit{et al.}~\cite{thakur2023benchmarking}, Chip-Chat~\cite{blocklove2023chip},
Chip-GPT~\cite{chang2023chipgpt}.  However, due to limited Verilog data sources, these works mainly target the scale of simple and small circuits (e.g., \textless20 designs with a medium of \textless45 HDL lines)~\cite{lu2023rtllm}. The relatively low scalability and solution quality have propelled the \textit{second trend} -- {enrich Verilog source}. Like oil, data is an immensely valuable resource. One could not generate high quality and comprehensive HDL codes without having LLMs trained on vast amount of such data. RTLLM~\cite{lu2023rtllm} and VerilogEval~\cite{liu2023verilogeval} introduce  specialized benchmarking framework (i.e., 30 designs from RTLLm and 156 designs from HDLBits~\cite{HDLBits} from VerilogEval) to assess the generation quality of LLMs. However, they either do not offer Power, Performance, and Area (PPA) analysis for the generated codes (e.g., VerilogEval), or the generated Verilog codes are directly extracted and synthesized using commercial tools to obtain PPA results, without considering PPA feedback (e.g., RTLLM). Thus, their solution quality is still limited.

\begin{figure*}
    \centering
    \includegraphics[width=0.98\textwidth]{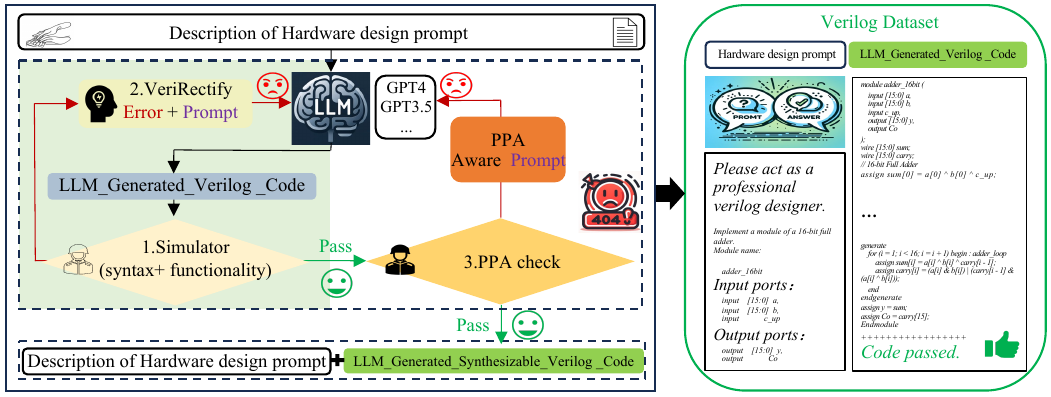} 
    \vspace{-4pt}
    \caption{This visualization captures the step-by-step process where an LLM synthesizes Verilog codes from hardware design prompts, with the ensuing code subjected to thorough validation by a Simulator and scrutinized for adherence to Power-Performance-Area (PPA) checks.
    }
    \label{figure:Code_generation}
\end{figure*}

In this work, as the first attempt, we integrate power, performance, and area-constraints into Verilog generation, and propose \framework, a open-source framework with multi-round Verilog generation and error feedback, 
 shown in Figure~\ref{figure:Code_generation}.  
We first generate initial Verilog codes using LLMs, followed by a unique two-stage refinement process. The first stage focuses on improving the syntax and functionality, while the second stage aims to optimize the code in line with PPA constraints, an essential aspect to ensure hardware design quality.  
Compared with state-of-the-arts (SOTAs), e.g., RTLLM~\cite{lu2023rtllm}, VerilogEval~\cite{liu2023verilogeval}, our \framework achieves a success rate of  62.0\% (+16\%) for functional accuracy and 81.37\% (+8.3\%) for syntactic correctness in Verilog code generation.
Our key contributions are summarized here:
\begin{itemize}

\item We use the detailed error diagnostics from the iverilog simulator~\cite{Williams2023}, and pinpoint the exact location of syntactic  or functional discrepancies as indicated by testbench failures as new prompts. We use multi-round generation to enhance the syntax and functionality correctness. 

\item To further ensure that the generated Verilog codes are \emph{synthesizable}, and design quality (PPA) is sound, we use Synopsys Design Compiler to
perform logic synthesis (and technology mapping) on the open source ASAP 7nm Predictive PDK, and check all designs' warnings/errors, and PPA report. We then integrate these PPA reports and warnings/errors with our PPA goal into the next round prompt for further refinement.

\item 
We incorporate in-context learning (ICL) to significantly improve the LLM performance in generating Verilog codes with only a few demonstration examples, especially when labeled data are scarce. By carefully selecting diverse text-to-Verilog pairs, ICL demonstrates superior performance and generalization capabilities compared to fine-tuning in limited example scenarios, thus increasing the performance of Verilog code generation.


\end{itemize}

\section{Background and Related Work}
There are mainly two directions related to Verilog generation.

\noindent \textbf{Finetune LLMs.}
Thakur \textit{et al.}~\cite{thakur2023benchmarking} advocate for the fine-tuning of open-source LLMs such as CodeGen~\cite{nijkamp2022codegen} to specifically generate Verilog code tailored for target designs. Subsequently, Chip-Chat~\cite{blocklove2023chip} delves into the intricacies of hardware design using LLMs,
highlighting the markedly superior performance of ChatGPT compared to other open-source LLMs.
Chip-GPT~\cite{chang2023chipgpt} also focuses on the task of RTL design by leveraging the capabilities of ChatGPT. 
These studies pave the way for a promising future where language models play a pivotal role in facilitating and enhancing various aspects of agile hardware design exploration. However, these works mainly target the scale of simple and small circuits (e.g., <20 designs with a medium of <45 HDL lines), as pointed out in~\cite{lu2023rtllm}.

\noindent \textbf{Enrich Verilog Source.} Like oil, data is an immensely valuable resource. Recent efforts have been focusing on enriching the Verilog data source.
RTLLM~\cite{lu2023rtllm}
introduces a benchmarking framework consisting of 30 designs that are specifically aimed at enhancing the scalability of benchmark designs. Furthermore, it utilizes effective prompt engineering techniques to improve the generation quality.
VerilogEval~\cite{liu2023verilogeval} assesses the performance of LLM in the realm of Verilog code generation for hardware design and verification. It comprises 156 problems from the Verilog instructional website HDLBits.  However, VerilogEval~\cite{liu2023verilogeval} does not offer PPA analysis for the generated codes. In RTLLM, the generated Verilog codes are directly extracted and synthesized using commercial tools to obtain PPA results, without PPA constraint-based feedback. 
Thus they suffer from limited generation quality.


\section{VeriPPA FRAMEWORK}
\subsection{Design Overview}

In our \framework framework, as illustrated in Figure~\ref{figure:Code_generation}, we use a text-based Description of Hardware Design, designated as $L$ and encapsulated within $design\_description.txt$ files, to serve as input/prompt for the LLMs. This description $L$ outlines the hardware design, details the module name, and specifies both input and output signals with the corresponding bit widths. This approach parallels traditional hardware design methodologies, where a designer would typically interpret $L$ to understand the overview of circuit requirements and then write the Verilog code, referred to as $V$. In a similar fashion, our framework's LLM is adept at parsing the $L$ description and subsequently writing the corresponding Verilog code, $V$. The synthesized code $V$ is then subjected to a rigorous validation sequence, beginning with a Simulator that checks both syntax and functionality. In the first loop, if unsuccessful, we will input the outcomes, along with any syntax and functionality errors, into the LLM for the generation of subsequent attempts. If successful, the code undergoes Power-Performance-Area (PPA) checks to ensure compliance with constraints. In the second loop, we check all designs’ warnings/errors  during  logic synthesis and
PPA reports. If not satisfied, both the design and its corresponding PPA report
will be fed back to the VeriRectify (Section~\ref{sec:VeriRectify}) for refinement.
This validation workflow ensures that the LLM-generated Verilog codes not only meets functional specifications but is also optimized for PPA considerations.


\subsection{Code Generation and Testing}
Utilizing design descriptions $L$, the LLM generates Verilog codes.
Our framework \framework incorporates the ICARUS Verilog simulator \cite{Williams2023} to automate the testing of the generated codes. In contrast to languages such as Python, Verilog requires the use of testbenches, denoted as \( T \), for comprehensive evaluation. These testbenches, \( T = \{T_1, T_2, \ldots, T_m\} \) are inputs to the ICARUS Verilog simulator \cite{Williams2023}.

that systematically assess the code's functionality, encompassing a wide array of test scenarios.

The integration of the ICARUS simulator into in \framework framework enhances the verification process, enabling an evaluation function 
$V$ (code,~$T$) that provides immediate feedback on the code's syntactical and operational integrity. This integrated approach contrasts with frameworks such as RTLLM \cite{lu2023rtllm}, where an external simulator is used to check the correctness of the generated Verilog codes. Since taking out the generated code can not provide instant detailed diagnostics of the code. Therefore, we need an integrated approach to get the detailed diagnostics of errors.
With \framework, detailed error reports from the simulator expedite the identification of faults, facilitating their rectification and refining the code generation process. This iterative, error-informed development cycle is discussed in further detail in the following sections, illustrating how precise error detection informs subsequent iterations of code generation within \framework framework.

 \begin{figure}[t]
    \centering
    \includegraphics[width=1\linewidth]{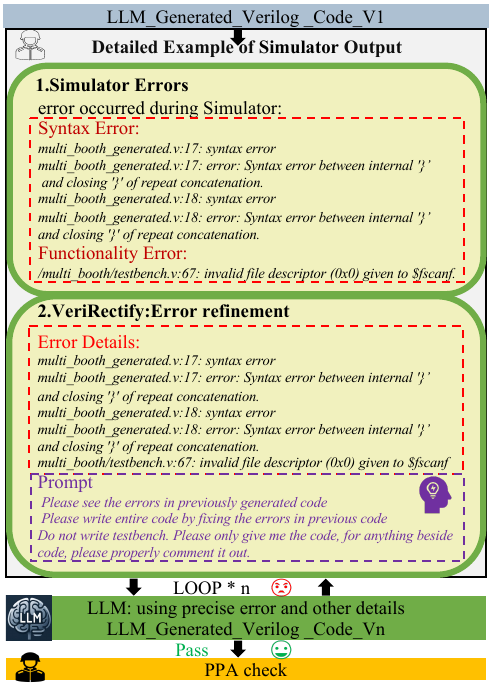} 
    \caption{The diagram illustrates the process of syntactic and functional code verification.
    }
    \label{fig:rectify}
\end{figure}

\subsection{VeriRectify}
\label{sec:VeriRectify}
In our approach, we leverage detailed error diagnostics from the simulator for generated codes. This refinement process, termed "\textit{VeriRectify}," is pivotal to ensuring the integrity of the generated Verilog codes. The VeriRectify phase ingests diagnostics from the iverilog simulator \cite{Williams2023}, pinpointing the exact location of syntactic discrepancies or functional discrepancies as indicated by testbench failures. 
This diagnostic data, when amalgamated with the antecedent code generation attempt
in the process where the LLM generates Verilog code.  The LLM then employs this enriched prompt to rectify and evolve the Verilog code toward compliance with specified correctness criteria, following the iterative relation \begin{equation}
V(n + 1) = V(n) - E(V(n))
\end{equation}
 where \( V(n) \) is the Verilog code at iteration \( n \) and \( E(V(n)) \) represents the identified errors. The VeriRectify workflow is depicted in Fig.~\ref{fig:rectify}. The figure’s top section displays the Syntax Error and Functional Error found in the output of our simulation tool (e.g., \( \text{booth\_multiplier} \)), which is the output from our simulation tool. During the rectification phase (Error Refinement), we utilize a tailored prompt (integrating the Error Details) for the LLMs, enabling it to correct the flaws identified in the Verilog codes initially produced.
 This approach allows the LLM to specifically address the encountered issues, which is fundamentally different from RTLLM~\cite{lu2023rtllm} (using a generalized prompt for all designs). As evident in Figure~\ref{fig:rectify}, the LLM effectively amends the issues in the Verilog codes for the \( \text{booth\_multiplier} \) design.
\subsection{Multi-round Conversation with Error Feedback}
To further refine the generated Verilog codes, we employ a multi-round conversation with error feedback loop analogous to human problem-solving techniques. This approach can be conceptualized as a function that iteratively refines the output by considering the errors of previous steps.
Let $V_i$ represent the Verilog code output after the $i^{th}$ iteration, and $E_i$
be the corresponding set of identified errors. Initially, 
$V_0$ is the first generated code with its errors $E_0$ The core of this approach is the refinement function, 
$R(V_i,E_i)$, which inputs the current code $V_i$ and its errors 
$E_i$, outputting an improved code version 
$V_{i+1}$. Concurrently, an error detection function $D(V_i)$ identifies errors in $V_i$, producing
$E_i$. The iterative process can be viewed as follows:\begin{equation} \label{eq:chain_of_thought}
  V_{i+1} = R(V_i, E_i) \quad \text{and} \quad E_{i+1} = D(V_{i+1})
\end{equation}



This process repeats until either no errors are detected or a predefined iteration limit, $K$ is reached, shown in Algorithm~\ref{alg:verilog_multi}. The iteration halts if, 
$D(V_{i+1})$=$\emptyset$ 
or 
$i=K$. 
$K$ is
empirically
adjustable based on observed diminishing returns in LLM performance improvements across iterations.
Thus, the multi-round conversation method in this context is a systematic, iterative algorithm aimed at progressively minimizing the error set
$E_i$ in the generated Verilog code 
$V_i$, enhancing code quality with each iteration until an optimal or satisfactory solution is reached within the bounds of 
$K$.

\begin{algorithm}[t]
\caption{Multi-Round Verilog Code Generation using LLM}
\label{alg:verilog_multi}
\small
\begin{algorithmic}[1]
\REQUIRE User prompt $P$, iteration limit $K$
\ENSURE Correct Verilog code $V_{final}$ or $V_K$
\STATE $i \gets 0$
\STATE $conv \gets []$ \# initialize a new conversation history list
\STATE $conv \gets \text{Append user prompt }(P) \text{ to conversation } $
\STATE $V_i \gets \text{LLM-generated code using prompt } P \text{in history} conv $
\STATE $E_i \gets \text{Error detection function using simulator}(D(V_i))$

\WHILE{$E_i \neq \emptyset$ AND $i < K$}
    \STATE $conv \gets \text{Append error information }(E_i) \text{ to conversation } $
    \STATE $V_{i+1} \gets \text{LLM-generated code using prompt } P_{new}$
    \STATE $E_{i+1} \gets \text{Error detection using simulator }(D(V_{i+1}))$
    \STATE $i \gets i + 1$
\ENDWHILE
\IF{$E_i = \emptyset$}
    \STATE $V_{final} \gets V_i$
\ELSE
    \STATE $V_{final} \gets V_K$
\ENDIF
\RETURN $V_{final}$
\end{algorithmic}
\end{algorithm}


\subsection{Power Performance \& Area (PPA) Checking }

The \emph{VeriRectify} process ensures the design to pass both RTL syntax check and cycle-accurate functional simulation. However, RTL simulation does not guarantee that the design (Verilog code) is \emph{synthesizable}. Furthermore, the quality of the hardware design must be measured by its power, performance, and area metrics.

Our approach takes a step further by inspecting PPA of the design $V$ which passes the \emph{VeriRectify} process as the following:
\begin{equation} \label{eq:PPA}
V = \left\{ \begin{matrix} 
V
& \text{if}~PPA(V)~\text{satisfies},
\\ 
VeriRectify(V, PPA(V))
& \text{otherwise.}
\end{matrix} \right. 
\end{equation}

In this work, our PPA check calls Synopsys Design Compiler to perform logic synthesis (and technology mapping) on the open-source ASAP 7nm Predictive PDK~\cite{vashishtha2017asap7}. We check all designs' warning/error messages during the logic synthesis, and the power ($nW$), area ($\mu m^2$), and clock frequency ($MHz$) for quality.
When the Verilog design can be synthesized and meets the PPA goal, it results in a pass. Otherwise, both the design and its corresponding PPA report will be fed back to the VeriRectify (Section~\ref{sec:VeriRectify}) for refinement. 

\begin{figure*}[t]
    \centering
    \includegraphics[width=0.99\textwidth]{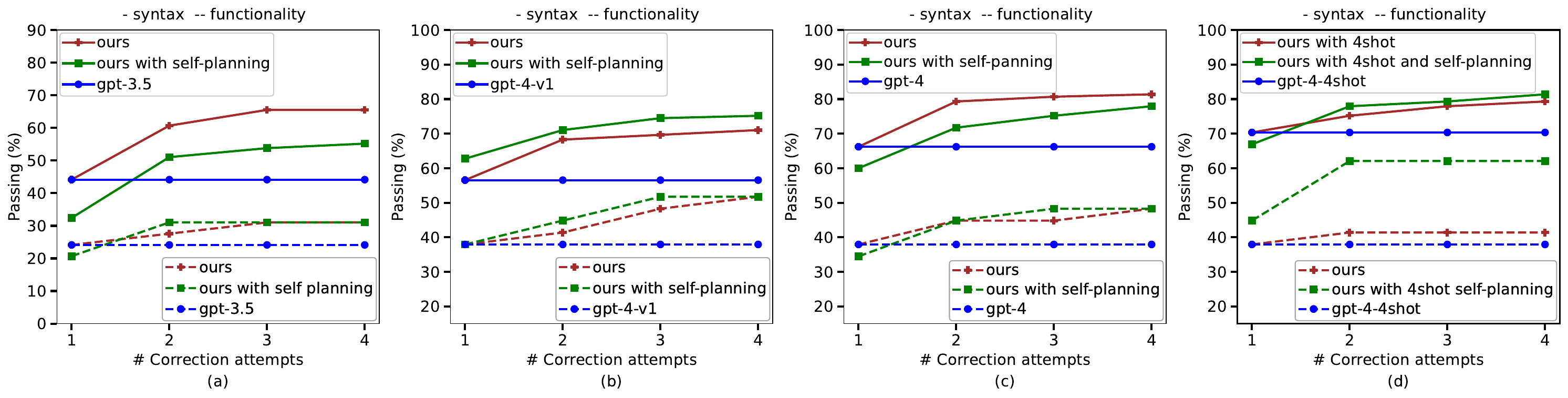} 
    \vspace{-4pt}
    \caption{ Correctness of generated Verilog code with respect to correction attempt on RTLLM, using (a) GPT-3.5; (b) GPT-4-v1; (c) GPT-4; (d) GPT-4-4shot}
    \vspace{-3pt}
    \label{fig:pass_rate}
\end{figure*}

\subsection{In-Context Learning}
LLMs have demonstrated remarkable in-context learning (ICL) capabilities. Given a few input-label pairs as demonstrations, they can predict labels for unseen inputs without requiring parameter updates~\cite{radford2019language, brown2020language}. In-context learning can be viewed as a meta-optimizer and may also be regarded as implicit fine-tuning~\cite{dai2023can}. While off-the-shelf LLM APIs do not offer extensive customization options, such as fine-tuning, ICL can significantly improve LLM's in-domain performance with only a few examples. In situations where labeled data is scarce, ICL has been shown to outperform explicit fine-tuning~\cite{perez2021true} on downstream tasks substantially. The availability of small training set poses a challenge in avoiding overfitting; however, it highlights the advantage of ICL over fine-tuning in scenarios with limited examples. In these cases, ICL demonstrates superior performance and better generalization capabilities, while fine-tuning may struggle with overfitting issues and result in poor generalization. A single LLM can achieve robust performance on multiple tasks using only its text interface: a few task examples are provided to the model as a prompt, accompanied by a query input, and the model generates a continuation to produce a predicted output for that query. 

\begin{equation}
    q(t | v) = \prod_{k=1}^K q(t_k | t_{< k}),
    \label{eq:icl}
\end{equation}

In this equation, $t_{k}$ represents the $k$-th language token of the input text, $t_{<k}$ denotes the set of preceding tokens, $v$ indicates demonstration example tokens, and $q$ is parameterized by an LLM. The equation describes the in-context learning of a large language model, where the model learns to predict the next token in the sequence by considering the previous demonstration examples.

The selection of few-shot examples~\cite{perez2021true, albalak2023improving} also influences the generalization capability of LLMs on downstream tasks. In our case, we carefully select the text-to-Verilog pairs to ensure that the examples cover a range of different Verilog designs, such as addition, multiplication, single-stage design, and pipelined design.




\begin{table}[t]
\begin{center}
\caption{\label{PPA_table} PPA results of generated Verilog code }
\resizebox{0.48\textwidth}{!}{
\begin{tabular}{|c| c | c | c | c | c |c|} 
 \hline 
 \textbf {Design Name} & \multicolumn{3}{c||}{\textbf{GPT-4}} & \multicolumn{3}{c||}{\textbf{GPT-4 (4-shot)}}\\ 
 \hline 
  & \multicolumn{1}{c|}{\begin{tabular}{@{}c@{}}Clock \\ (ps)\end{tabular}} & \multicolumn{1}{c|}{\begin{tabular}{@{}c@{}}Power \\ (\(\scriptstyle\mu\)W)\end{tabular}} & \multicolumn{1}{c||}{\begin{tabular}{@{}c@{}}Area \\ (\(\scriptstyle\mu\)m\(^2\))\end{tabular}} & \multicolumn{1}{c|}{\begin{tabular}{@{}c@{}}Clock \\ (ps)\end{tabular}} & \multicolumn{1}{c|}{\begin{tabular}{@{}c@{}}Power \\ (\(\scriptstyle\mu\)W)\end{tabular}} & \multicolumn{1}{c||}{\begin{tabular}{@{}c@{}}Area \\ (\(\scriptstyle\mu\)m\(^2\))\end{tabular}}  \\
 \hline \hline
 adder\_8bit & 318.5 & 6.3 & 38.5 & 333.1 & 6.1 & 42.9 \\
 \hline
 adder\_16bit & 342.2 & 10.9 & 104.5 & 135.1 & 41.1 & 152.8 \\
 \hline
 adder\_32bit & 500.0 & 14.2 & 211.6 & 500.0 & 14.7 & 213.2 \\
 \hline
 multi\_booth & 409.0 & 112.1 & 526.0 & 409.0 & 112.1 & 526.0 \\
 \hline
 right\_shifter & 47.5 & 144.3 & 42.9 & 47.5 & 144.3 & 42.9 \\
 \hline
 width\_8to16 & 74.1 & 223.2 & 145.8 & 145.6 & 128.7 & 157.2 \\
 \hline
 edge\_detect & 61.5 & 49.0 & 23.3 & 61.5 & 49.0 & 23.3 \\
 \hline
 mux & 54.7 & 215.3 & 86.1 & 54.7 & 215.3 & 86.1 \\
 \hline
 pe & 500.0 & 552.5 & 2546.5 & 500.0 & 541.0 & 2488.6 \\
 \hline
 asyn\_fifo & 295.2 & 406.4 & 1279.3 & 228.3 & 526.6 & 1295.4 \\
 \hline
 counter\_12 & 134.4 & 33.1 & 40.6 & 124.5 & 34.6 & 36.4 \\
 \hline
 fsm & 88.3 & 32.7 & 31.5 & 68.7 & 49.0 & 50.2 \\
 \hline
 multi\_pipe\_4bit & 254.7 & 40.7 & 131.3 & - & - & - \\
 \hline
 pulse\_detect & 10.3 & 187.5 & 13.5 & 32.7 & 59.1 & 13.5 \\
 \hline
 calendar & - & - & - & 208.6 & 86.6 & 199.0 \\
 \hline
\end{tabular}}
\end{center}
\end{table}

\section{Evaluation}
\subsection{Datasets}
In assessing our \framework framework, we utilize two benchmark datasets. Firstly, the RTLLLM dataset\cite{lu2023rtllm} includes 29 designs. Notably, it originally contained 30 designs, but the risc\_cpu  design is currently unavailable. Secondly, we employ the VerilogEval dataset \cite{liu2023verilogeval}, which comprises two subsets: VerilogEval-human, featuring 156 designs, and VerilogEval-machine, consisting of 108 designs.

\subsection{Experimental Setup}
We demonstrate the effectiveness of our \framework framework for generating PPA-optimized Verilog code for the given designs. We adopt, GPT-3.5 \cite{openai_gpt3_5} and GPT-4 \cite{openai_gpt4} as our LLM models. We use n=1, temperature temp = 0.7, and a context length of 2048 in our setting. Further, we incorporate the ICARUS Verilog simulator \cite{Williams2023} to automate the testing of the generated code. 
For PPA check, we perform the logic synthesis using Synopsys Design Compiler with \texttt{compile\_ultra} command and we use the ASAP 7nm Predictive PDK~\cite{vashishtha2017asap7}. 
We implemented a flow (Python script) that sweeps the timing constraints to find the fastest achievable clock frequency for all the generated designs. 
All experiments were conducted on a Linux-
based host with AMD EPYC 7543 32-Core Processor and an
NVIDIA A100-SXM 80 GB.

\subsection{Generation Correctness}
This study evaluates Verilog code generation accuracy using two primary metrics: syntax checking and functionality verification. Figure \ref{fig:pass_rate} presents our methodology for improving Verilog correctness through successive correction attempts and self-planing~\cite{lu2023rtllm}. We generate five different codes for each design description, attempting up to four corrections within each generation. However, after certain attempts, the efficiency of these corrections diminishes, as the LLMs tend to provide repetitive responses to identical errors.

In Figure \ref{fig:pass_rate}, we plot syntax and functionality correctness percentages against the number of correction attempts. The graph features a solid line for syntax correctness and a dotted line for functionality correctness. Functionality is evaluated the same as RTLLM \cite{lu2023rtllm}, considering a design functionally correct if at least one generated code passes the functionality test. We use GPT-3.5 and observe initial syntax correctness of 44.13\% and functionality correctness of 24.13\%, as shown in Figure \ref{fig:pass_rate} (a). After applying correction attempts, these figures improved to 65.51\% for syntax and 31.03\% for functionality. Compared to RTLLM, which initially scored 24.82\% in syntax and 27.58\% in functionality without corrections. 
 Please note that RTTLM uses self panning in prompt. 
 Integrating our correction approach with RTLLM increases the maximum syntax and functionality correctness to 49.65\% and 34.48\%, respectively.

We then evaluate two versions of the GPT-4 model. The first, GPT-4-0314 (v1), showed an initial syntax correctness of 56.55\% and functionality correctness of 37.93\%. Our correction methods raised these to 71.03\% for syntax and 51.72\% for functionality. Combining RTLLM with our approach, we further enhanced the syntax correctness to 75.17\% and functionality to 51.72\%, as indicated in Figure \ref{fig:pass_rate} (b).
For the base GPT-4 model, we noticed an increase in syntax correctness from 66.2\% to 81.37\% by the fourth attempt and in functionality from 37.93\% to 48.27\%. When combined with RTLLM and our correction techniques, the syntax correctness further improved from 60\% to 77.93\%, and functionality from 34.48\% to 48.27\%, as shown in Figure \ref{fig:pass_rate} (c).

Finally, testing the GPT-4 model with four-shot learning, we observed an improvement in syntax from 70.34\% to 79.31\% and in functionality from 37.93\% to 41.37\%. With the addition of RTLLM and our correction methods, the functionality correctness notably increased from 44.82\% to 62.06\% after four attempts, as demonstrated in Figure \ref{fig:pass_rate} (d). This significant improvement highlights the effectiveness of our methods in enhancing the functional accuracy of the designs.

\begin{figure}[ht]
    \centering
    \includegraphics[width=0.45\textwidth]{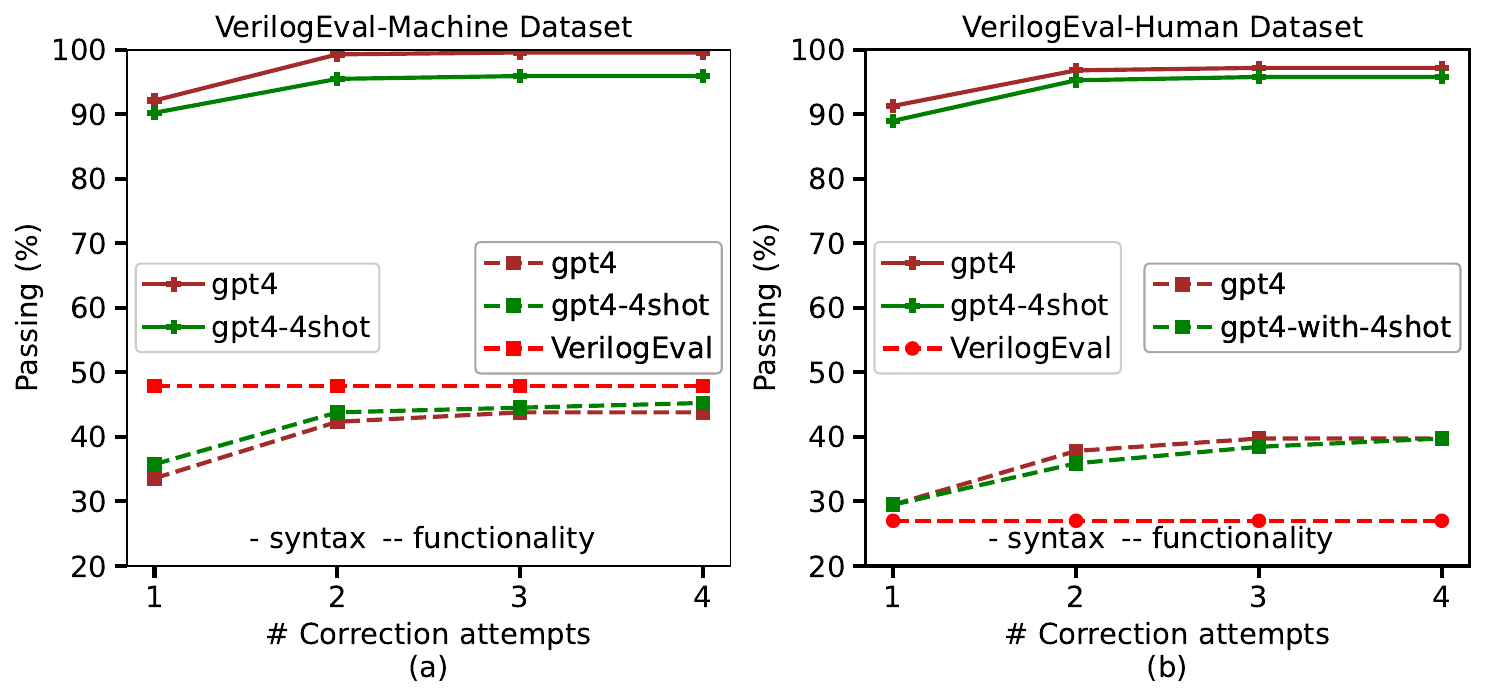} 
    \caption{Correctness of generated Verilog code with respect to correction attempt on VerilogEval, using (a) VerilogEval-Machine, and (b) VerilogEval-Human.}
    \label{fig:pass_rate_nvidia}
    \vspace{-2mm}
\end{figure}

In evaluating our \framework framework with VerilogEval data, we found notable improvements. For the VerilogEval-Machine dataset (Figure \ref{fig:pass_rate_nvidia} (a)), our method significantly increased syntax accuracy. Functionality accuracy also rose from 33.57\% to 43.79\% using GPT-4, and further to 45.25\% with GPT-4’s four-shot learning. The VerilogEval-human dataset showed similar trends, with functionality accuracy improving from 29.48\% to 39.74\% through the application of GPT-4 and its four-shot learning variant as shown in the Figure \ref{fig:pass_rate_nvidia} (b). This underscores our framework's effectiveness in enhancing both syntax and functionality in Verilog code generation

\begin{figure}[t]
    \centering
      \includegraphics[width=1\linewidth]{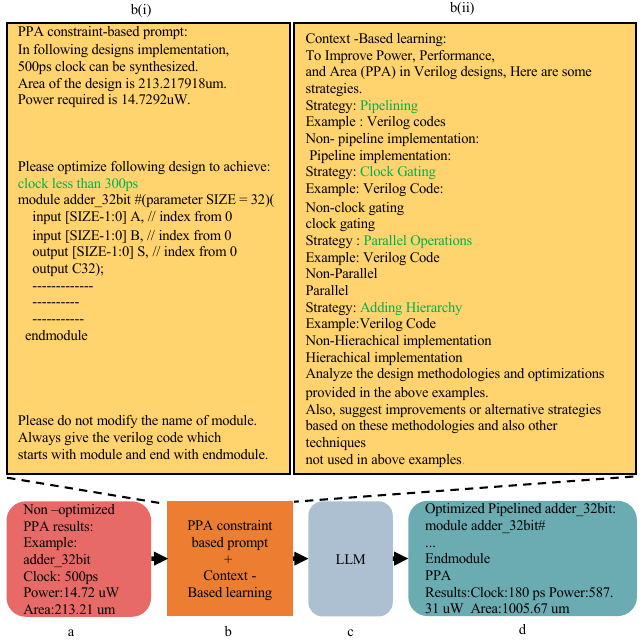}
  \captionof{figure}{Optimization Flow.}
    \label{fig:optimization_flow}
\end{figure}

\begin{table}[t]
\begin{center}
\caption{\label{PPA_table_optimized} Optimized results}
\resizebox{0.8\linewidth}{!}{
\begin{tabular}{|c||c|c|c|} 
 \hline 
 \textbf{Design Name} & \textbf{Clock (ps)} & \textbf{Power (\(\mu\)W)} & \textbf{Area (\(\mu\)m)}   \\
 \hline \hline
 adder\_32bit & 180.0 & 587.31 & 1005.67 \\
 \hline
 multi\_booth  & 123.2 & 42.39 & 42.92 \\
 \hline
 pe  & 325.0 & 1206.0 & 4863.88 \\
 \hline
 asyn\_fifo &  114.8 & 988.92 & 1344.86 \\
 \hline
 radix2\_div & - & - & - \\
 \hline
\end{tabular}}
\end{center}
\end{table}

\subsection{PPA Optimization}
In  \framework, we use the Synopsis Design Compiler for synthesizing the designs, culminating in the production of PPA reports. The PPA results of complex designs are encapsulated in Table \ref{PPA_table}. This table, though comprehensive, does not encompass specific design constraints. Similar to the ChipGPT approach \cite{chang2023chipgpt}, where an output manager and enumerative search finalize the PPA from multiple reports, our process also generates multiple PPA reports for each design. An example is the $pulse\_detect$ design, which consistently met criteria across five evaluations of passing functionally and syntactically. Therefore, in post-synthesis, we collated five PPA reports for the $pulse\_detect$ design, and we selected the most optimized one to include in Table \ref{PPA_table}.

It is crucial that PPA results do not conform to specialized design requirements, a standard practice in industrial applications. 
To address this disparity, 
we further perform the PPA constraint-based feedback mechanism, integrated with context-based learning, as illustrated in Figure \ref{fig:optimization_flow}. 
This approach represents a significant step towards aligning LLM-generated code with industry-specific PPA requirements.
Figure \ref{fig:optimization_flow} demonstrates our process, starting with the collection of synthesized design outputs that require optimization. 
For example, $adder\_32bit$, is initially synthesized with a 500ps clock as shown in Figure \ref{fig:optimization_flow} (a). 
To enhance the speed of $adder\_32bit$, we impose a clock constraint, aiming for a clock speed of less than 300ps, as outlined in the PPA constraint-based prompt in Figure \ref{fig:optimization_flow}. The framework instructs the LLM to consider various optimization strategies, including Pipelining, Clock Gating, Parallel Operation, and Hierarchical Design. It also encourages the exploration of additional methods to generate Verilog code that meets the defined optimization constraints, as illustrated in the context-based learning segment of Figure \ref{fig:optimization_flow}.

Upon providing the PPA-based constraint prompt and context to the LLM, we analyze the resultant Verilog code for syntax and functional accuracy, making corrections where necessary. If the code passes both checks, we proceed to its final synthesis, achieving an optimized Verilog code as shown in Figure \ref{fig:optimization_flow} (d), where the $adder\_32bit$ operates at an improved 180ps clock. In Table \ref{PPA_table_optimized}, we present the results of selected optimized designs. Due to the page limit and we only show substantial optimizations, the table does not include simpler designs. Notably, no design from the VerilogEval \cite{liu2023verilogeval} dataset features in Table \ref{PPA_table_optimized}, as those designs did not require complex optimization.

\section{conlcusion}

In this paper, we introduce a novel framework \framework, designed to assess and enhance LLM efficiency in this specialized area. Our method includes generating initial Verilog code using LLMs, followed by a unique two-stage refinement process. The first stage focuses on improving the functional and syntactic integrity of the code, while the second stage aims to optimize the code in line with Power-Performance-Area (PPA) constraints, an essential aspect of effective hardware design.
This dual-phase approach of error correction and PPA optimization has led to notable improvements in the quality of LLM-generated Verilog code. Our framework schieves 62.0\% (+16\%) for functional accuracy and 81.37\% (+8.3\%) for syntactic correctness in Verilog code generation, compared to SOTAs.
\bibliographystyle{plain}

\bibliography{bib/ref}
\end{document}